\documentclass{article}

\usepackage{PRIMEarxiv}

\usepackage[utf8]{inputenc} 
\usepackage[T1]{fontenc}    
\usepackage{hyperref}       
\usepackage{url}            
\usepackage{booktabs}       
\usepackage{amsfonts}       
\usepackage{nicefrac}       
\usepackage{microtype}      
\usepackage{lipsum}
\usepackage{fancyhdr}       
\usepackage{graphicx}       
\graphicspath{{media/}}     
\usepackage{cite}
\usepackage{amsmath,amssymb,amsfonts}
\usepackage{textcomp}
\usepackage{algorithmicx}
\usepackage{algorithm}
\usepackage[noend]{algpseudocode}
\usepackage{multirow}
\usepackage{array}
\usepackage{float}
\usepackage{caption}
\usepackage{subcaption}
\title{FAM: FAST ADAPTIVE FEDERATED META-LEARNING
}

\author{
  Indrajeet Kumar Sinha \\
  Department of Information Technology\\
  Indian Institute of Information Technology Allahabad\\
  Prayagraj, Uttar Pradesh, India. \\
  \texttt{pcl2016004@iiita.ac.in} \\
   \And
  Shekhar Verma \\
  Department of Information Technology\\
  Indian Institute of Information Technology Allahabad\\
  Prayagraj, Uttar Pradesh, India. \\
  \texttt{sverma@iiita.ac.in} \\
  \AND
  Krishna Pratap Singh \\
  Department of Information Technology\\
  Indian Institute of Information Technology Allahabad\\
  Prayagraj, Uttar Pradesh, India. \\
  \texttt{kpsingh@iiita.ac.in} \\
}

\begin{document}
\maketitle

\begin{abstract}
In this work, we propose a fast adaptive federated meta-learning (FAM) framework for collaboratively learning a single global model, which can then be personalized locally on individual clients. Federated learning enables multiple clients to collaborate to train a model without sharing data. Clients with insufficient data or data diversity participate in federated learning to learn a model with superior performance. Nonetheless, learning suffers when data distributions diverge. There is a need to learn a global model that can be adapted using clients’ specific information to create personalized models on clients is required. MRI data suffers from this problem, wherein, one, due to data acquisition challenges, local data at a site is sufficient for training an accurate model and two, there is a restriction of data sharing due to privacy concerns and three, there is a need for personalization of a learnt shared global model on account of domain shift across client sites. The global model is sparse and captures the common features in the MRI. This skeleton network is grown on each client to train a personalized model by learning additional client-specific parameters from local data. Experimental results show that the personalization process at each client quickly converges using a limited number of epochs. The personalized client models outperformed the locally trained models, demonstrating the efficacy of the FAM mechanism. Additionally, the sparse parameter set to be communicated during federated learning drastically reduced communication overhead, which makes the scheme viable for networks with limited resources.
\end{abstract}


\section{Introduction}
Deep learning \cite{DeepL} models be very good at capturing intricate relationships in MRI data due to their complex patterns and representations learning ability. Training with a larger dataset is usually more stable for deep networks, even though data quality, diversity, and relevance are also crucial. For a high signal-to-noise ratio, scanning takes a long time involving the usage of multiple saturation frequencies and multiple acquisitions. Since dynamic data acquisition with high spatial resolution and temporal rate is challenging, fast image acquisition leads to undersampling. Accelerated MRI with undersampled acquisitions enhances efficiency by recovering missing data using reconstruction algorithms incorporating prior information \cite{Sparse-MRI,MR-parameter}. Deep reconstruction models are trained to conditionally map undersampled acquisitions to fully sampled images \cite{Prior-Guided,feng2021donet,Deep-Cascade}. In \cite{guo2021multi}, Adversarial alignment between source and target sites is suggested to improve the similarity of latent-space representations. Reconstruction models in \cite{feng2022specificity} used a global encoder and local decoders trained separately at each site. Both methods use conditional models, assuming imaging operator knowledge, but outperform across-site variability. Conditional models fail to generalize when the imaging operator changes \cite{biswas2019dynamic}. The training and test sets must match acceleration rates and sampling densities. These may restrict multi-institutional collaboration \cite{knoll2019assessment}. In \cite{elmas2022federated}, FedGIMP: cross-site learning of a generative MRI prior and prior adaptation following the global MRI prior is learned via an unconditional adversarial model to synthesize high-quality MR images. Classification models also suffer due to low signal-to-noise ratio and undersampling. Thus, the need for training of generalizable models on large and diverse MRI datasets is not met as sufficient data is not available. This prevents learning diverse descriptions of health disorders and datasets at a single institution. FL \cite{federated_taxonomy} addresses this limitation of limited data size in one institution and allows multiple institutions to collaborate to train a global model while adhering to medical data privacy regulations.\\

In FL, decentralized models collaborate without sharing confidential medical records. A centralized server obtains parameters from models trained on local data by clients. It then aggregates the parameters from locally trained models to compute a set of global model parameters. These parameters are broadcast to the local clients, where they form the parameters of the local model and are trained on locally stored private data at every client. The trained parameters or gradients are sent back to the server for aggregation. This iterative process of collaborative learning without data sharing continues to obtain a final model. However, datasets on different clients may show significant diversity in MRI distribution (variation in scanners) and accelerated imaging operators (variation in acceleration rates and sample densities). This leads to a domain shift between clients. Significant domain shifts across sites or training and test sets can suffer from performance declines for multisite models \cite{knoll2019assessment,liu2021feddg}. The performance of Vanilla federated \cite{federated_learning_survey_1,federated_healthcare} classification models may be impaired where significant data heterogeneity exists across sites. In such a scenario of undersampled images, limited data on each client, non-sharing of local data and suboptimal global FL model, we need to adapt the global model into a personalized model using the data at the client. The FAM mechanism attempts to fulfil these requirements. This work demonstrates sparse meta-learner training that yields good generalizable initialization in federated architecture that quickly adapts to new tasks. 

The significant contributions are outlined as follows:
\begin{itemize}
    \item We propose a fast adaptive federated meta-learning model for MRI image classification. FAM yields a sparse initialization model with a small message footprint. It learns better generalized global representation and is able to adapt quickly to the local domain-specific information on individual clients.

    \item The model is implemented on multiple MRIs like $Brain-MRI$ \cite{BMRI}, $Cardiac-MRI$ \cite{Cardiac}, $Alzheimer-MRI$ \cite{Alzheimer}, $Breast Cancer-MRI$ \cite{Breast}, and $Prostate-MRI$ \cite{Prostate} image dataset.
    
\end{itemize}



\section{Problem Description}

The Vanilla FL mechanism endeavours to obtain a global model that has higher accuracy as compared to the models trained by nodes using their local data. This is based on the assumption that nodes’ data come from the same underlying distribution. However, the global model is suboptimal when the data from different nodes are heterogeneous. MRI data is characterized by domain shift between clients that results in data heterogeneity across nodes. Getting personalized models for each individual node, while exploiting the data from other nodes is a problem that needs to be addressed. Coupled with the problem of personalization is the problem of huge computation and communication overhead that ails the nodes. It can be observed that a common global sparse model can be learnt at the server through parameter aggregation, while nodes learn quick in-situ adaptation using their local data. Learning a sparse model in the FL mechanism would require a small number of parameters, and personalization can be achieved using meta-learning \cite{hospedales2021meta}. The FL meta-learning mechanism would result in a sparse global model using minimal computation and communication overhead followed by meta-learning-based personalization of the sparse global model at each node using their local data.

\section{Fast Adaptive Federated Meta-learning (FAM) Mechanism}
 
FAM mechanism is based on the observations that there is insufficient data for classification at one place with data available with different nodes that can be utilized for learning but cannot be shared due to privacy concerns and regulations. Moreover, there are domain shifts in data across clients and a global model that is learnt by vanilla FL mechanism is not suitable for clients. The shared global model needs to be personalized locally at each client. The conjecture is that the shared global model must learn the main common features that can be used as good initialization for personalization on the client nodes. A global model that has captured the homogeneity across clients should be able to learn and adapt to the specific domain shift with limited local data at a particular client. To capture the homogeneity, such a global model must be sparse. During federated learning, the number of parameters to be computed at the clients and communicated to the server would be small, leading to low computation and communication overheads. This warrants that a pruning strategy is adopted to learn a sparse global model during federated learning. After the learnt global model is shared with the clients, it needs to be personalized at each client node. This domain shift adaptation can be achieved using meta-learning. 
The FAM mechanism tries to address the low client-specific performance when the global shared model is vulnerable to domain shifts due to data heterogeneity and suffers from low sensitivity to client-specific features, especially when the model is not adapted for cross-client variability at client nodes. The federated learning mechanism utilizes a sparsification strategy using Lottery Ticket Hypothesis (LTH) strategy \cite{LTH2018} and MAML-based meta-learning \cite{MAML-model} that adapts locally by growing the sparse global model. 
\vspace{-0.5em}        

\begin{figure*}[!h]
    \centering
    \includegraphics[trim=21 140 102 40, clip, scale=0.52]{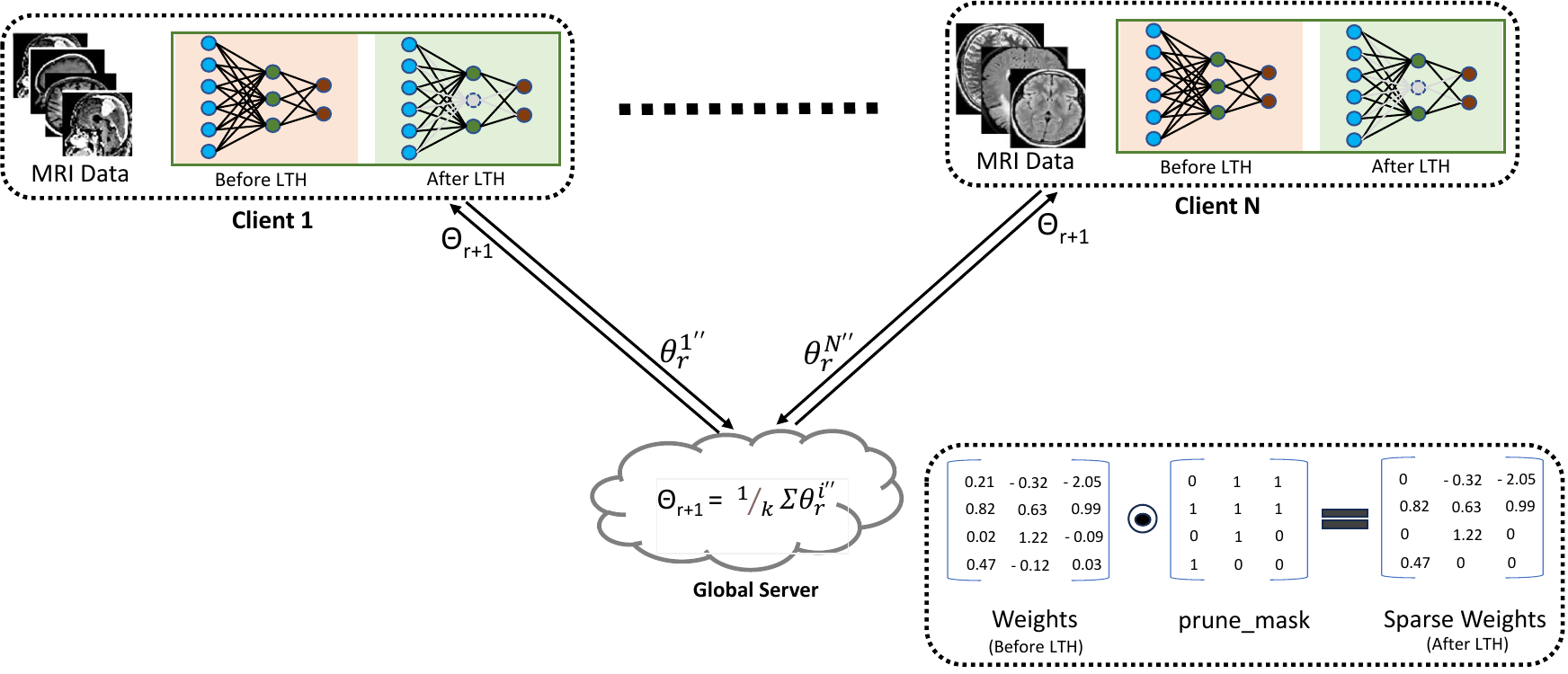}  
    \caption{\small{Overview of the FAM mechanism. The neural network is initially dense; as the training progresses, the connections between some neurons are dropped, also called sparsifying. After retraining some rounds, disconnected links between neurons are established, also called growing. The network is then reset to the dense state and retrained to give the final parameters.}}
\end{figure*}

\subsection{FAM mechanism}
FAM mechanism endeavours to train a global model using federated learning. The global model is sparse and learns common features from multiple MRI features present in data across different client sites. To learn the common features, the lottery ticket hypothesis is used as a sparsification strategy to prune the model during the federated learning process. In the first phase of federated meta-learning, multiple clients train machine learning models using their local data. The model parameters are sent by clients and aggregated at the server.
            \begin{equation}\label{eq:FL}
                \Theta^* = \underset{\Theta}{ \operatorname{argmin}} \hspace{0.6em} \sum_{i=1}^{k}\frac{m_i}{M} {\mathbf{l}_i}(\Theta)
            \end{equation}
            Here, $\mathbf{l}_{i}$(·) expresses the loss of global model parameters $\Theta$ on local data at $i^{th}$ client.
        The meta-learning uses Model-Agnostic Meta-Learning (MAML) to learn parameter initialization that facilitates rapid adaptation to new tasks. To achieve this, it solves the meta-objective (equation \ref{eq:ML}) by utilizing meta-training dataset $D^{tr}$ with randomly initialized parameters $\theta$ in order to find the optimal meta-parameter $\theta^{*}$.
        \begin{equation}\label{eq:ML}
                \theta^* = \underset{\theta}{ \operatorname{argmax}} \hspace{0.6em} \log p(\Theta|D^{tr})
        \end{equation} 
        MAML consists of two phases: meta-training and meta-testing. During the meta-training phase, the model is exposed to a distribution of tasks. The model is denoted by the parametric function $f_{\theta}$ with parameters $\theta$. As the model adapts to a new task $T_i$, its parameters become $\theta_{T_i}^{'}$. This updated parameter $\theta_{T_i}^{'}$ is calculated by applying one or more gradient descent updates to task $T_i$. Mathematically expressed as
        \begin{equation}\label{eq:MAML-1}
                \theta_{T_i}^{'} = \theta -   \alpha {\nabla}_{\theta}\mathcal{L}_{T_i} (f_{\theta}(D^{tr}_{i}))
        \end{equation}
        where $\alpha$ is the learning rate. The meta-optimization is performed on the model parameters $\theta$, whereas the loss is computed using the updated task-specific model parameters $\theta_{T_i}^{'}$.
        \begin{equation}\label{eq:MAML-2}
                \theta^{''} = \theta -   \beta \hspace{0.1em}{\nabla}_{\theta}{\sum_{{T_i}\sim p(T)}}\mathcal{L}_{T_i} (f_{\theta_{T_i}^{'}}(D^{ts}_{i}))
        \end{equation}
        where $\beta$ is the meta step size.
        

\floatname{algorithm}{Algorithm}
\renewcommand{\algorithmicensure}{\textbf{Outcome:}}
\newcommand{\End}{\textbf{END}}
\begin{algorithm}[!h]
    \caption{Federated Sparse Meta-training:}\label{alg:Meta-Client Federated}
    \begin{algorithmic}[1]
    \footnotesize
        \Require \hspace{-0.2em}$p(\mathcal{T}^{\hspace{0.1em}i})$: distr. of tasks for \textit{$i^{th}$\hspace{0.1em}Client};\hspace{0.1em} $\alpha$,\hspace{0.0em} $\beta$, $prune\_mask$
        \Ensure Updated \textit{$i^{th}$\hspace{0.1em}Client} Parameters: ${\theta_{r}^{\hspace{0.1em}i^{\hspace{0.1em}''}}}$ 
    
        \hspace{-3em}\textbf{for} r in rounds \textbf{do}
        
        \hspace{-2.0em}\textbf{for} i in clients \textbf{do}
        \State \textit{{Initialise:} local clients with Server parameters.}
        
            \hspace{-1.2em} ${\theta_{r}^{\hspace{0.1em}i} \leftarrow \Theta_{r}}$ 
        \raggedright 
        
        \State \textit{{Inner Loop:} for a batch of tasks $\mathcal{T}_{j}^{\hspace{0.1em}i} \sim  p(\mathcal{T}^{\hspace{0.1em}i})$ for $i^{th}$\hspace{0.0em}Client}. \\ 
            \hspace{1em}\textbf{for all} $\mathcal{T}_{j}^{\hspace{0.1em}i}$ \textbf{do} \\
            \hspace{2.5em}${\theta_{j,\hspace{0.1em}r}^{\hspace{0.1em}i^{\hspace{0.1em}'}}} \leftarrow \theta_{r}^{\hspace{0.1em}i} - \alpha {\nabla_{\theta}}\mathcal{L}_{\mathcal{T}_{j}^{\hspace{0.1em}i\hspace{0.1em}}}(f_{\theta^{\hspace{0.1em}i}_{r}}) $ \\ 
        \raggedright
            \hspace{1em}\textbf{end for}

        \State \textit{{Outer Loop:} optimize over average of all task losses.} \\
        \hspace{1em}  {\textbf{if} $flag < 2$ {\textbf{then}} }\\ \hspace{1.5em}
            \hspace{1em} ${\theta_{r}^{\hspace{0.1em}i^{\hspace{0.1em}''}}} \leftarrow \theta_{r}^{\hspace{0.1em}i} - \beta \hspace{0.1em}{\nabla_{\theta}} \sum_{\mathcal{T}_{j}^{\hspace{0.1em}i} \sim  p(\mathcal{T}^{\hspace{0.1em}i})}\mathcal{L}_{\mathcal{T}_{j}^{\hspace{0.1em}i\hspace{0.1em}}}(f_{\theta_{r}^{\hspace{0.1em}i^{\hspace{0.1em}'}}}) $ \\ 
        \raggedright
        \hspace{1em}  {\textbf{else}  }\\
                    \hspace{3em} ${\theta_{r}^{\hspace{0.1em}i^{\hspace{0.1em}''}}} \leftarrow \theta_{r}^{\hspace{0.1em}i} - \beta \hspace{0.1em}{\nabla_{\theta}} \sum_{\mathcal{T}_{j}^{\hspace{0.1em}i} \sim  p(\mathcal{T}^{\hspace{0.1em}i})}\mathcal{L}_{\mathcal{T}_{j}^{\hspace{0.1em}i\hspace{0.1em}}}(f_{\theta_{r}^{\hspace{0.1em}i^{\hspace{0.1em}'}}})$ \\ \vspace{0.2em}
                    \hspace{6.5em} $\odot prune\_mask $ 
        \raggedright
        
        \State  \textit{{Client shares parameters to Server.} } 
        
        \hspace{-0em}\textbf{end for} \\
        \hspace{-1em}\textbf{end for} \\
    \end{algorithmic}
\end{algorithm}

Each round of client-side federated meta-training is described in Algorithm \ref{alg:Meta-Client Federated}. All clients are initially configured with the same Server parameters ($W_r$) for $r = 0$. Then, clients begin training the model using their local batch of tasks ($\mathcal{T}_{j}^{\hspace{0.1em}i}$). Each client sends its updated parameters to the server after training the meta-learner over several tasks, and then the server aggregates $k$ clients updates from the $n$-client set. We chose the value of $k = 2$. By aggregating $k$-selected parameters, the server computes global parameters. We articulate client-side training and server-side aggregation as one-round communication.

\begin{algorithm}[!h]
    \footnotesize
    \caption{Sparse training server-side}\label{alg:server Federated}
    \begin{algorithmic}[1]
        \Require k, $flag = 0$; \textit{Client\hspace{2px}i\hspace{1px}}: $\theta_{r}^{\hspace{0.1em}i^{\hspace{0.1em}''}}$
        \Ensure Global Parameters: ${\Theta_{}^{*}}$ \newline
        \hspace{-3em}//Server selects k-client parameters from $clients\_pool$.
        \State \textit{\textbf{Aggregation:}}  \newline   
            \hspace{1.5em}    ${\Theta_{r+1} \leftarrow \frac{1}{k} \sum_{i}^{i=k} \theta_{r}^{\hspace{0.1em}{i^{{\hspace{0.1em}''}}}}}$
 
        \raggedright
        \State \textit{\textbf{Condition for LTH:}} \newline
            \hspace{0.5em}  {\textbf{if} $r ==$ ``LTH" {\textbf{then}} }\\ \hspace{1.5em} {\textit{``Go for LTH"}}\\
            \hspace{1.5em} {\hspace{0.5em}$flag = 1$}\\
        \State \textit{\textbf{Sparsification:}} \newline    
            \hspace{0.5em}  {\textbf{if} $flag == 0$  {\textbf{then}} }\\ \hspace{1.5em} {\textit{No sparsification}}\newline
            \hspace{0.5em}  {\textbf{elif} $flag == 1$  {\textbf{then}} }\\ \hspace{1.5em} {\textit{``Performing LTH": ${\Theta_{r+1} \leftarrow \Theta_{0}} \odot prune\_mask$}} \\
            \hspace{1.5em} \textit{``Broadcast prune\_mask to Clients"} \\
            \hspace{1.5em} {\hspace{0.5em}$flag = 2$}\newline    
            \hspace{0.5em}  {\textbf{elif} $flag == 2$  {\textbf{then}} }\\ \hspace{1.5em} {\textit{``Sparsification": ${\Theta_{r+1} \leftarrow \Theta_{r+1}} \odot prune\_mask$}}
        \State \textbf{Broadcast to Clients:} ${\Theta_{r+1}}$ 
    \end{algorithmic}
\end{algorithm}

After one round of the client's meta-training, the server performs, described in Algorithm \ref{alg:server Federated}, differently before and after pruning in which weights with $p\%$ of the smallest magnitude, and the remaining weights are reset with the original initial weights and then updated weights $W_{r+1}$ and prune\_mask communicated to all clients. The server's prune\_mask preserves sparsity at the client's training. Hereafter, client-server communication updates the local and global models using the same server's prune\_mask.

    We drop the $p$ connections in the neural network once throughout the training stage. This $p$ is calculated as mentioned in the equation \ref{eq:Prune}.

    \begin{equation}\label{eq:Prune}
        p = (|\Theta| -||\Theta ||_{0}) + prune\_rate * ||\Theta||_{0} \\
    \end{equation}
            

\begin{algorithm}[h]
\footnotesize
    \caption{sparsification}\label{alg:prune}
    \begin{algorithmic}[1]
        \Require { \textit{weight, prune\_rate}}
        \Ensure {prune\_mask}
        \State \textit{ // Count zeroes in aggregated weights at the server.} \newline
             ${num\_zeros \leftarrow (weight==0).sum()}$ \newline
             ${num\_nonzeros \leftarrow weight.count() - num\_zeros}$
        
        \State \textit{ // Number of parameters to be sparsed.} \newline
            ${num\_prune \leftarrow prune\_rate * num\_nonzeros}$ 
        \State \textbf{if} {num\_remove == 0:}
        
        \hspace{-1.4em} {\textbf{return} weight.bool()}
        \State {$p \leftarrow num\_zeros + num\_prune$}
        \State {$w, id_w \leftarrow sort(weight)$}
        \State {$prune\_mask \leftarrow weight.bool().int()$}
        \State {$prune\_mask.data.view(-1)[id_w[:p]] \leftarrow 0 $}
        \State {\textbf{return} prune\_mask} \\
        \End
    \end{algorithmic}
\end{algorithm}

In Algorithm \ref{alg:prune}, the procedure for sparsification is depicted. Sparsifying parameters is performed by changing $1$ to $0$ in a binary mask that turns off the gradient flow. In the binary mask, $1$ indicates the active weights, while $0$ indicates inactive weights. The prune\_rate is a hyperparameter that determines the number of parameters to be sparsed. The prune\_mask with dropped weights is computed.

This is achieved by growing the connections in the neural network. Each connection is associated with weight. These weights are updated through the training process (i.e., a gradient is calculated for the loss function concerning each weight in the network). We froze the previously learned weights for old connections (i.e., sparsed global meta-learned parameters), and then adaptation or personalization is performed by updating disconnected connections (i.e., weights with 0 value).

\begin{algorithm}[h]
\footnotesize
    \caption{Model personalization on growing connections:}\label{alg:Meta Personalisation}
    \begin{algorithmic}[1]
        \Require $\mathcal{T}_{new}^{\hspace{0.1em}i}:$ samples for new task for \textit{$i^{th}$\hspace{0.1em}Client};\hspace{0.1em} ${\Theta_{}^{*}}$, $\alpha$
        \Ensure Personalized Parameters: ${\theta^{\hspace{0.1em}i^{\hspace{0.1em}*}}}$ 

         \State \hspace{-0.5em} \textbf{for} $epoch$ in epochs \textbf{do} 
         
         \State \hspace{0.3em} // \textit{\textbf{Client's parameter initialization.}} 
         
            \hspace{-1.7em} ${\theta_{}^{\hspace{0.1em}i} \leftarrow \Theta_{}^{*}}$ \hfill
        \raggedright 
        \State \hspace{0.3em} // \textit{\textbf{Personalizing $i^{th}$client on samples of \hspace{0.2em}$\mathcal{T}_{new}^{\hspace{0.1em}i}$.}} 
        
            \hspace{0.3em} ${\theta_{}^{\hspace{0.1em}i^{\hspace{0.1em}*}}} \leftarrow \theta_{}^{\hspace{0.1em}i} - \alpha {\nabla}_{\theta_{}^{\hspace{0.1em}i}}\mathcal{L}(\mathcal{T}_{new}^{\hspace{0.1em}i},\theta_{}^{\hspace{0.1em}i}) \odot inverse(prune\_mask) $ \\ 
        
        \raggedright
        \vspace{0.2em}

        \State \hspace{-0.5em} \textbf{end for} 
        
    \end{algorithmic}
\end{algorithm}

Personalization of the model is performed for task-specific samples. Algorithm \ref{alg:Meta Personalisation} illustrates the procedure for model personalization. Parameters update over the training is controlled by setting initial non-zero parameters to unchanged, and the rest are grown. This is achieved by growing the connections in the neural network. Each connection is associated with weight. These weights are updated through the training process (i.e., a gradient is calculated for the loss function concerning each weight in the network). We froze the previously learned weights for old connections (i.e., sparsed global meta-learned parameters), and then adaptation or personalization is performed by updating disconnected connections (i.e., weights with 0 value). Growing the parameters works on changing 0 to 1 in a binary mask, i.e. we say inverse(prune\_mask). Changing 0 to 1 in a binary mask enables the gradient flow. Weights pruned in the sparsification step are activated in the personalization stage.

\section{Results and Discussion}
\subsection{Dataset Description}
We demonstrate FAM's performance on the following datasets: $CIFAR-100$, $MiniImagenet$, $Brain-MRI$, $Cardiac-MRI$, $Alzheimer-MRI$, $Breast Cancer-MRI$, and $Prostate-MRI$. 

The $MiniImagenet$ dataset contains 50000 training and 10000 testing images distributed evenly across 100 classes. Each image is $84\times84$ pixels in size and has three channels. The $CIFAR-100$ dataset is a subset of the Tiny Images dataset that contains images from $100$ classes, each with $500$ training images and $100$ testing images; each sample is an RGB $32\times32$ pixels image. 

The $Brain-MRI$ dataset contains $3264$ images of human brain MRI that are used to categorise the brain into four categories: glioma, meningioma, no tumour, and pituitary. We combined the glioma, meningioma, and pituitary tumour classes into a single tumour class for our needs. The dataset is divided into $500$ images for the Healthy (no tumour) class and $2764$ for the Sick (tumour) class.

The $Cardiac-MRI$ dataset contains $62421$ cardiac MRI images for Coronary artery disease, divided into $36630$ normal and $25791$ sick classes. 

The $Alzheimer-MRI$ dataset includes $6400$ MRI images of Alzheimer's disease. Previously, the dataset contained four image classes. Three different levels of Demented classes were combined into a single Demented class. As a result, the dataset is evenly distributed between the Healthy (Non-Demented) and Sick (Demented) classes. Each image was $128\times128$ pixels in size and had three channels. 

The dataset $Breast Cancer-MRI$ contains $1480$ breast MRI images for breast cancer disease that are evenly distributed between the Healthy (Benign) and Sick (Malignant) classes. Each image was $250\times540$ pixels in size for three channels. 

The $Prostate-MRI$ dataset consists of $1528$ prostate MRI images in the transverse plane that are evenly distributed into two classes: significant and nonsignificant. Each image was $384\times384$ pixels in size for three channels. 

To ensure consistency throughout meta-training, we resized all MRI images to an RGB $120\times120$ pixel size. 

\subsection{Model Setup}


\subsubsection{Meta-Learner Model Setting}
We used the same architecture for the meta-learning model as \cite{MAML-model}, which has four modules with $3\times3$ convolutions and $64$ filters, followed by batch normalization, a ReLU nonlinearity and $2\times2$ max-pooling. The dimensionality of the last hidden layer is $1600$ for $MiniImagenet$, $256$ for $CIFAR-100$, and $3136$ for the simulated healthcare datasets. The loss function for all models is the cross-entropy error between the predicted and true classes.
\subsubsection{Federated training}
We created 20 clients with the same meta-learner model architecture for simulation; each client has a local dataset. 
Each client trains the model over their local data and shares it with the global server. The server performs two main events: one is the selection ok $k$-clients, and the second is the aggregation of selected $k$-client's parameters to produce generalized global parameters. Here, the server is responsible for applying LTH \cite{malach2020proving} to sparsify the network parameters.

\subsection{Eperimental Results}
Our FAM mechanism were trained on $CIFAR-100$, $MiniImagenet$, and five MRI datasets, and the best results were reported in the table above. For $CIFAR-100$ and $MiniImagenet$, we used an internal learning rate of 0.01, a meta-learning rate of 0.0001, ten local epochs, ten tasks, ten clients, and 6000 communication rounds, while for simulated healthcare data, we used 1000 rounds. We used the LTH  after 50\% of the training rounds to refine our model. Until the final iteration, we kept the sparsity at 80\% on $CIFAR-100$ and $MiniImagenet$ and 70\% on simulated healthcare data.

We aim to learn the minimum common feature out of all data to which one can quickly adapt.

\begin{table}[!h]
    \caption{\small{FAM mechanism performance on $CIFAR-100$.}}`
    \label{table:FAM-Cifar}
    \centering
    \scriptsize
    \begin{tabular}{|@{}c@{}|c||c|c|c|c|}
    \hline
        \multicolumn{6}{|c|}{FAM - $CIFAR-100$} \\ \hline
        \textbf{S. No.\hspace{0.1em}}  & \textbf{Adaptation} & \textbf{Accuracy} & \textbf{Precision} & \textbf{Recall} & \textbf{F1-score}\\ \hline
        
        1 & 5 way - 0 shot & 20.00 & 4.00 & 20.00 & 6.67 \\ \hline
        \multicolumn{6}{|c|}{\#1 Episode} \\ \hline
        2 & 5 way - 1 shot & 32.00 & 30.67 & 32.00 & 30.38 \\ \hline
        3 & 5 way - 5 shot & 40.00 & 31.31 & 40.00 & 34.23 \\ \hline
        \multicolumn{6}{|c|}{\#5 Episode} \\ \hline
        4 & 5 way - 1 shot & 32.80 & 35.09 & 32.80 & 28.60 \\ \hline
        5 & 5 way - 5 shot & 49.60 & 59.14 & 49.60 & 45.25 \\ \hline
        
    \end{tabular}
\end{table}

\textbf{Table \ref{table:meta-cifar}} shows the FAM mechanism results on five classes of $Cifar-100$. We achieved accuracy, precision, recall and f1-score of 20.00\%, 4.00\%, 20.00\% and 6.67\% for 5 way - 0 shot on without adaption and after adaption achieved a maximum for five episodes of 5 way - 5 shot of 49.60\%, 59.14\%, 49.60\% and 45.25\%, respectively.

\begin{table}[!h]
    \caption{\small{FAM mechanism performance on $MiniImagenet$.}}
    \label{table:FAM-Mini}
    \centering
    \scriptsize
    \begin{tabular}{|@{}c@{}|c||c|c|c|c|}
    \hline
        \multicolumn{6}{|c|}{FAM - $MiniImagenet$} \\ \hline
        \textbf{S. No.\hspace{0.1em}}  & \textbf{Adaptation} & \textbf{Accuracy} & \textbf{Precision} & \textbf{Recall} & \textbf{F1-score}\\ \hline
        
        1 & 5 way - 0 shot & 20.00 & 4.00 & 20.00 & 6.67 \\ \hline
        \multicolumn{6}{|c|}{\#1 Episode} \\ \hline
        2 & 5 way - 1 shot & 36.00 & 33.33 & 36.00 & 32.66 \\ \hline
        3 & 5 way - 5 shot & 52.00 & 47.46 & 52.00 & 48.10 \\ \hline
        \multicolumn{6}{|c|}{\#5 Episode} \\ \hline
        4 & 5 way - 1 shot & 44.00 & 33.81 & 44.00 & 37.31 \\ \hline
        5 & 5 way - 5 shot & 64.00 & 72.89 & 64.00 & 64.63 \\ \hline
    \end{tabular}
\end{table}

\textbf{Table \ref{table:FAM-Mini}} shows the FAM mechanism results on five classes of $MiniImagenet$. We achieved accuracy, precision, recall and f1-score similar to the $CIFAR-100$ dataset for 5 way - 0 shot on without adaption and after adaption achieved a maximum for five episodes of 5 way - 5 shot of 64.00\%, 72.89\%, 64.00\% and 64.63\%, respectively.

\begin{table}[!h]
\scriptsize
    \caption{ \small{FAM mechanism performance on $Brain$-$MRI$. }} 
    \label{table:FAM-Brain}
    \centering
    \begin{tabular}{|@{}c@{}|c||c|c|c|c|}
    \hline
        \multicolumn{6}{|c|}{FAM - $Brain$-$MRI$} \\ \hline
        \textbf{S. No.\hspace{0.1em}}  & \textbf{Adaptation} & \textbf{Accuracy} & \textbf{Precision} & \textbf{Recall} & \textbf{F1-score}\\ \hline
        
        1 & 2 way - 0 shot & 72.00 & 75.28 & 72.00 & 71.06 \\ \hline
        \multicolumn{6}{|c|}{\#1 Episode} \\ \hline
        2 & 2 way - 1 shot & 75.00 & 77.47 & 75.00 & 74.42 \\ \hline
        3 & 2 way - 5 shot & 81.00 & 81.31 & 81.00 & 80.95 \\ \hline
        \multicolumn{6}{|c|}{\#5 Episode} \\ \hline
        4 & 2 way - 1 shot & 82.00 & 82.84 & 82.00 & 81.88 \\ \hline
        5 & 2 way - 5 shot & 89.00 & 89.02 & 89.00 & 89.00 \\ \hline
        
    \end{tabular}
\end{table}
\textbf{Table \ref{table:FAM-Brain}} shows the FAM mechanism results on $Brain$-$MRI$. We achieved accuracy, precision, recall and f1-score of 72.00\%, 75.28\%, 72.00\% and 71.06\% for 5 way - 0 shot on without adaption and after adaption achieved a maximum for five episodes of 5 way - 5 shot of 89.00\%, 89.02\%, 89.00\% and 89.00\%, respectively.

\begin{table}[!h]
\scriptsize
    \caption{ \small{FAM mechanism performance on $Cardiac$-$MRI$. }} 
    \label{table:FAM-Cardiac}
    \centering
    \begin{tabular}{|@{}c@{}|c||c|c|c|c|}
    \hline
        \multicolumn{6}{|c|}{FAM - $Cardiac$-$MRI$} \\ \hline
        \textbf{S. No.\hspace{0.1em}}  & \textbf{Adaptation} & \textbf{Accuracy} & \textbf{Precision} & \textbf{Recall} & \textbf{F1-score}\\ \hline
        
        1 & 2 way - 0 shot & 55.00 & 55.49 & 55.00 & 53.96 \\ \hline
        \multicolumn{6}{|c|}{\#1 Episode} \\ \hline
        2 & 2 way - 1 shot & 56.00 & 56.68 & 56.00 & 54.84 \\ \hline
        3 & 2 way - 5 shot & 66.00 & 66.03 & 66.00 & 65.99 \\ \hline
        \multicolumn{6}{|c|}{\#5 Episode} \\ \hline
        4 & 2 way - 1 shot & 62.00 & 62.02 & 62.00 & 61.98 \\ \hline
        5 & 2 way - 5 shot & 69.00 & 69.01 & 69.00 & 69.00 \\ \hline
        
    \end{tabular}
\end{table}

\textbf{Table \ref{table:FAM-Cardiac}} shows the FAM mechanism results on $Cardiac$-$MRI$. We achieved accuracy, precision, recall and f1-score of 55.00\%, 55.49\%, 55.00\% and 53.96\% for 5 way - 0 shot on without adaption and after adaption achieved a maximum for five episodes of 5 way - 5 shot of 69.00\%, 69.01\%, 69.00\% and 69.00\%, respectively.

\begin{table}[!h]
\scriptsize
    \caption{ \small{FAM mechanism performance on $Alzheimer$-$MRI$. }} 
    \label{table:FAM-Alz}
    \centering
    \begin{tabular}{|@{}c@{}|c||c|c|c|c|}
    \hline
        \multicolumn{6}{|c|}{FAM - $Alzheimer$-$MRI$} \\ \hline
        \textbf{S. No.\hspace{0.1em}}  & \textbf{Adaptation} & \textbf{Accuracy} & \textbf{Precision} & \textbf{Recall} & \textbf{F1-score}\\ \hline
        
        1 & 2 way - 0 shot & 62.00 & 66.45 & 62.00 & 59.24 \\ \hline
        \multicolumn{6}{|c|}{\#1 Episode} \\ \hline
        2 & 2 way - 1 shot & 68.00 & 68.03 & 68.00 & 67.99 \\ \hline
        3 & 2 way - 5 shot & 70.00 & 72.98 & 70.00 & 69.00 \\ \hline
        \multicolumn{6}{|c|}{\#5 Episode} \\ \hline
        4 & 2 way - 1 shot & 73.00 & 73.23 & 73.00 & 72.93 \\ \hline
        5 & 2 way - 5 shot & 77.00 & 77.01 & 77.00 & 77.00 \\ \hline
        
    \end{tabular}
\end{table}

\textbf{Table \ref{table:FAM-Alz}} shows the FAM mechanism results on $Alzheimer$-$MRI$. We achieved accuracy, precision, recall and f1-score of 55.00\%, 55.49\%, 55.00\% and 53.96\% for 5 way - 0 shot on without adaption and after adaption achieved a maximum for five episodes of 5 way - 5 shot of 77.00\%, 77.01\%, 77.00\% and 77.00\%, respectively.

\begin{table}[!h]
\scriptsize
    \caption{ \small{FAM mechanism performance on $BreastCancer$-$MRI$. }} 
    \label{table:FAM-Breast}
    \centering
    \begin{tabular}{|@{}c@{}|c||c|c|c|c|}
    \hline
        \multicolumn{6}{|c|}{FAM - $BreastCancer$-$MRI$} \\ \hline
        \textbf{S. No.\hspace{0.1em}}  & \textbf{Adaptation} & \textbf{Accuracy} & \textbf{Precision} & \textbf{Recall} & \textbf{F1-score}\\ \hline
        
        1 & 2 way - 0 shot & 67.00 & 75.62 & 67.00 & 63.97 \\ \hline
        \multicolumn{6}{|c|}{\#1 Episode} \\ \hline
        2 & 2 way - 1 shot & 69.00 & 74.10 & 69.00 & 67.27 \\ \hline
        3 & 2 way - 5 shot & 75.00 & 75.50 & 75.00 & 74.88 \\ \hline
        \multicolumn{6}{|c|}{\#5 Episode} \\ \hline
        4 & 2 way - 1 shot & 70.00 & 73.81 & 70.00 & 68.75 \\ \hline
        5 & 2 way - 5 shot & 81.00 & 82.04 & 81.00 & 80.84 \\ \hline
        
    \end{tabular}
\end{table}

\textbf{Table \ref{table:FAM-Breast}} shows the FAM mechanism results on $BreastCancer$-$MRI$. We achieved accuracy, precision, recall and f1-score of 67.00\%, 75.62\%, 67.00\% and 63.97\% for 5 way - 0 shot on without adaption and after adaption achieved a maximum for five episodes of 5 way - 5 shot of 81.00\%, 82.04\%, 81.00\% and 80.84\%, respectively.

\begin{table}[!h]
\scriptsize
    \caption{ \small{FAM mechanism performance on $Prostate$-$MRI$. }} 
    \label{table:FAM-Prost}
    \centering
    \begin{tabular}{|@{}c@{}|c||c|c|c|c|}
    \hline
        \multicolumn{6}{|c|}{FAM - $Prostate$-$MRI$} \\ \hline
        \textbf{S. No.\hspace{0.1em}}  & \textbf{Adaptation} & \textbf{Accuracy} & \textbf{Precision} & \textbf{Recall} & \textbf{F1-score}\\ \hline
        
        1 & 2 way - 0 shot & 60.00 & 68.60 & 60.00 & 54.77 \\ \hline
        \multicolumn{6}{|c|}{\#1 Episode} \\ \hline
        2 & 2 way - 1 shot & 68.00 & 70.05 & 68.00 & 67.16 \\ \hline
        3 & 2 way - 5 shot & 74.00 & 75.00 & 74.00 & 73.74 \\ \hline
        \multicolumn{6}{|c|}{\#5 Episode} \\ \hline
        4 & 2 way - 1 shot & 72.00 & 72.92 & 72.00 & 71.72 \\ \hline
        5 & 2 way - 5 shot & 75.00 & 75.01 & 75.00 & 75.00 \\ \hline
        
    \end{tabular}
\end{table}

\textbf{Table \ref{table:FAM-Prost}} shows the FAM mechanism results on $Prostate$-$MRI$. We achieved accuracy, precision, recall and f1-score of 60.00\%, 68.60\%, 60.00\% and 54.77\% for 5 way - 0 shot on without adaption and after adaption achieved a maximum for five episodes of 5 way - 5 shot of 75.00\%, 75.01\%, 75.00\% and 75.00\%, respectively.


\subsection{Ablation Study}

In order to investigate the effect of the ablation, we evaluated both the network's performance separately. We compared the weights needed to train for adaptation in both cases.


\begin{table}[!h]
\scriptsize
    \caption{ \small{Federated meta-model performance on $CIFAR-100$.}} \label{table:meta-cifar}
    \centering
    \begin{tabular}{|@{}c@{}|c||c|c|c|c|}
    \hline
        \multicolumn{6}{|c|}{Federated Meta-Model - $CIFAR-100$} \\ \hline
        \textbf{S. No.\hspace{0.1em}}  & \textbf{Adaptation } & \textbf{Accuracy} & \textbf{Precision} & \textbf{Recall} & \textbf{F1-score}\\ \hline
        
        1 & 5 way - 0 shot & 20.00 & 4.00 & 20.00 & 6.67 \\ \hline
        \multicolumn{6}{|c|}{\#1 Episode} \\ \hline
        2 & 5 way - 1 shot & 28.00 & 22.74 & 28.00 & 24.23 \\ \hline
        3 & 5 way - 5 shot & 36.00 & 29.33 & 36.00 & 31.84 \\ \hline
        \multicolumn{6}{|c|}{\#5 Episode} \\ \hline
        4 & 5 way - 1 shot & 31.20 & 24.44 & 31.20 & 26.62 \\ \hline
        5 & 5 way - 5 shot & 40.00 & 32.34 & 40.00 & 35.56 \\ \hline
        
    \end{tabular}
\end{table}

\textbf{Table \ref{table:meta-cifar}} shows the adapted federated meta-learned model results on five classes of $Cifar-100$. We achieved accuracy, precision, recall and f1-score of 20.00\%, 4.00\%, 20.00\% and 6.67\% for 5 way - 0 shot on without adaption and after adaption achieved a maximum for five episodes of 5 way - 5 shot of 40.00\%, 32.34\%, 40.00\% and 35.56\%, respectively.

\begin{table}[!h]
\scriptsize
    \caption{ \small{Federated meta-model performance on $MiniImagenet$. }} \label{table:meta-mini}
    \centering
    \begin{tabular}{|@{}c@{}|c||c|c|c|c|}
    \hline
        \multicolumn{6}{|c|}{Federated Meta-Model - $MiniImagenet$} \\ \hline
        \textbf{S. No.\hspace{0.1em}}  & \textbf{Adaptation } & \textbf{Accuracy} & \textbf{Precision} & \textbf{Recall} & \textbf{F1-score}\\ \hline
        
        1 & 5 way - 0 shot & 20.00 & 4.00 & 20.00 & 6.67 \\ \hline
        \multicolumn{6}{|c|}{\#1 Episode} \\ \hline
        2 & 5 way - 1 shot &32.00 & 28.33 & 32.00 & 27.30 \\ \hline
        3 & 5 way - 5 shot & 48.00 & 46.16 & 48.00 & 42.14 \\ \hline
        \multicolumn{6}{|c|}{\#5 Episode} \\ \hline
        4 & 5 way - 1 shot & 40.00 & 30.50 & 40.00 & 33.98 \\ \hline
        5 & 5 way - 5 shot & 60.00 & 63.93 & 60.00 & 59.39 \\ \hline
        
    \end{tabular}
\end{table}

\textbf{Table \ref{table:meta-mini}} shows the adapted federated meta-learned model results on five classes of $MiniImagenet$. We achieved accuracy, precision, recall and f1-score similar to the $CIFAR-100$ dataset for 5 way - 0 shot on without adaption and after adaption achieved a maximum for five episodes of 5 way - 5 shot of 60.00\%, 63.93\%, 60.00\% and 59.39\%, respectively.

\begin{table}[!h]
\scriptsize
    \caption{ \small{Federated meta-model performance on $Brain$-$MRI$. }} \label{table:meta-brain}
    \centering
    \begin{tabular}{|@{}c@{}|c||c|c|c|c|}
    \hline
        \multicolumn{6}{|c|}{Federated Meta-Model - $Brain$-$MRI$} \\ \hline
        \textbf{S. No.\hspace{0.1em}}  & \textbf{Adaptation } & \textbf{Accuracy} & \textbf{Precision} & \textbf{Recall} & \textbf{F1-score}\\ \hline
        
        1 & 2 way - 0 shot & 73.00 & 74.17 & 73.00 & 72.67 \\ \hline
        \multicolumn{6}{|c|}{\#1 Episode} \\ \hline
        2 & 2 way - 1 shot & 74.00 & 77.57 & 74.00 & 73.13 \\ \hline
        3 & 2 way - 5 shot & 79.00 & 79.10 & 79.00 & 78.98 \\ \hline
        \multicolumn{6}{|c|}{\#5 Episode} \\ \hline
        4 & 2 way - 1 shot & 78.00 & 82.17 & 78.00 & 77.26 \\ \hline
        5 & 2 way - 5 shot & 85.00 & 85.01 & 85.00 & 85.00 \\ \hline
        
    \end{tabular}
\end{table}

\textbf{Table \ref{table:meta-brain}} shows the adapted federated meta-learned model results on $Brain$-$MRI$. We achieved accuracy, precision, recall and f1-score of 73.00\%, 74.17\%, 73.00\% and 72.67\% for 5 way - 0 shot on without adaption and after adaption achieved a maximum for five episodes of 5 way - 5 shot of 85.00\%, 85.01\%, 85.00\% and 85.00\%, respectively.

\begin{table}[!h]
\scriptsize
    \caption{ \small{Federated meta-model performance on $Cardiac$-$MRI$. }} \label{table:meta-cardiac}
    \centering
    \begin{tabular}{|@{}c@{}|c||c|c|c|c|}
    \hline
        \multicolumn{6}{|c|}{Federated Meta-Model - $Cardiac$-$MRI$} \\ \hline
        \textbf{S. No.\hspace{0.1em}}  & \textbf{Adaptation } & \textbf{Accuracy} & \textbf{Precision} & \textbf{Recall} & \textbf{F1-score}\\ \hline
        
        1 & 2 way - 0 shot & 53.00 & 53.15 & 53.00 & 52.42 \\ \hline
        \multicolumn{6}{|c|}{\#1 Episode} \\ \hline
        2 & 2 way - 1 shot & 55.00 & 55.25 & 55.00 & 54.45 \\ \hline
        3 & 2 way - 5 shot & 63.00 & 63.13 & 63.00 & 62.91 \\ \hline
        \multicolumn{6}{|c|}{\#5 Episode} \\ \hline
        4 & 2 way - 1 shot & 59.00 & 59.30 & 59.00 & 58.67 \\ \hline
        5 & 2 way - 5 shot & 68.00 & 68.26 & 68.00 & 67.88 \\ \hline
        
    \end{tabular}
\end{table}

\textbf{Table \ref{table:meta-cardiac}} shows the adapted federated meta-learned model results on $Cardiac$-$MRI$. We achieved accuracy, precision, recall and f1-score of 53.00\%, 53.15\%, 53.00\% and 52.42\% for 5 way - 0 shot on without adaption and after adaption achieved a maximum for five episodes of 5 way - 5 shot of 68.00\%, 68.26\%, 68.00\% and 67.88\%, respectively.

\begin{table}[!h]
\scriptsize
    \caption{ \small{Federated meta-model performance on $Alzheimer$-$MRI$. }} \label{table:meta-alz}
    \centering
    \begin{tabular}{|@{}c@{}|c||c|c|c|c|}
    \hline
        \multicolumn{6}{|c|}{Federated Meta-Model - $Alzheimer$-$MRI$} \\ \hline
        \textbf{S. No.\hspace{0.1em}}  & \textbf{Adaptation } & \textbf{Accuracy} & \textbf{Precision} & \textbf{Recall} & \textbf{F1-score}\\ \hline
        
        1 & 2 way - 0 shot & 60.00 & 64.57 & 60.00 & 56.60 \\ \hline
        \multicolumn{6}{|c|}{\#1 Episode} \\ \hline
        2 & 2 way - 1 shot & 62.00 & 63.79 & 62.00 & 60.73 \\ \hline
        3 & 2 way - 5 shot & 65.00 & 71.17 & 65.00 & 62.25 \\ \hline
        \multicolumn{6}{|c|}{\#5 Episode} \\ \hline
        4 & 2 way - 1 shot & 70.00 & 70.53 & 70.00 & 69.81 \\ \hline
        5 & 2 way - 5 shot & 75.00 & 75.09 & 75.00 & 74.98 \\ \hline
        
    \end{tabular}
\end{table}

\textbf{Table \ref{table:meta-alz}} shows the adapted federated meta-learned model results on $Alzheimer$-$MRI$. We achieved accuracy, precision, recall and f1-score of 60.00\%, 64.57\%, 60.00\% and 56.60\% for 5 way - 0 shot on without adaption and after adaption achieved a maximum for five episodes of 5 way - 5 shot of 75.00\%, 75.09\%, 75.00\% and 74.98\%, respectively.

\begin{table}[!h]
\scriptsize
    \caption{ \small{Federated meta-model performance on $BreastCancer$-$MRI$. }} \label{table:meta-breast}
    \centering
    \begin{tabular}{|@{}c@{}|c||c|c|c|c|}
    \hline

        \multicolumn{6}{|c|}{Federated Meta-Model - $BreastCancer$-$MRI$} \\ \hline 
        
        \textbf{S. No.\hspace{0.1em}}  & \textbf{Adaptation } & \textbf{Accuracy} & \textbf{Precision} & \textbf{Recall} & \textbf{F1-score}\\ \hline
        
        1 & 2 way - 0 shot & 63.00 & 65.78 & 63.00 & 61.29 \\ \hline
        \multicolumn{6}{|c|}{\#1 Episode} \\ \hline
        2 & 2 way - 1 shot & 62.00 & 64.88 & 62.00 & 60.07 \\ \hline
        3 & 2 way - 5 shot & 68.00 & 68.47 & 68.00 & 67.79 \\ \hline
        \multicolumn{6}{|c|}{\#5 Episode} \\ \hline
        4 & 2 way - 1 shot & 65.00 & 65.76 & 65.00 & 64.57 \\ \hline
        5 & 2 way - 5 shot & 79.00 & 79.29 & 79.00 & 78.95 \\ \hline
        
    \end{tabular}
\end{table}

\textbf{Table \ref{table:meta-breast}} shows the adapted federated meta-learned model results on $BreastCancer$-$MRI$. We achieved accuracy, precision, recall and f1-score of 63.00\%, 65.78\%, 63.00\% and 61.29\% for 5 way - 0 shot on without adaption and after adaption achieved a maximum for five episodes of 5 way - 5 shot of 79.00\%, 79.29\%, 79.00\% and 78.95\%, respectively.

\begin{table}[!h]
\scriptsize
    \caption{\small{Federated meta-model performance on $Prostate$-$MRI$. }} \label{table:meta-prost}
    \centering
    \begin{tabular}{|@{}c@{}|c||c|c|c|c|} 
    \hline
        \multicolumn{6}{|c|}{Federated Meta-Model - $Prostate$-$MRI$} \\ \hline
        \textbf{S. No.\hspace{0.1em}}  & \textbf{Adaptation } & \textbf{Accuracy} & \textbf{Precision} & \textbf{Recall} & \textbf{F1-score}\\ \hline
        
        1 & 2 way - 0 shot & 61.00 & 71.57 & 61.00 & 55.56 \\ \hline
        \multicolumn{6}{|c|}{\#1 Episode} \\ \hline
        2 & 2 way - 1 shot & 64.00 & 64.09 & 64.00 & 63.94 \\ \hline
        3 & 2 way - 5 shot & 72.00 & 73.87 & 72.00 & 71.44 \\ \hline
        \multicolumn{6}{|c|}{\#5 Episode} \\ \hline
        4 & 2 way - 1 shot & 69.00 & 70.38 & 69.00 & 68.47 \\ \hline
        5 & 2 way - 5 shot & 79.00 & 79.58 & 79.00 & 78.90 \\ \hline
        
    \end{tabular}
\end{table}

\textbf{Table \ref{table:meta-prost}} shows the adapted federated meta-learned model results on $Prostate$-$MRI$. We achieved accuracy, precision, recall and f1-score of 61.00\%, 71.57\%, 61.00\% and 55.56\% for 5 way - 0 shot on without adaption and after adaption achieved a maximum for five episodes of 5 way - 5 shot of 79.00\%, 79.58\%, 79.00\% and 78.90\%, respectively.

\begin{table}[!h]
\scriptsize
    \caption{ \small{Comparision for the number of non-zero parameters between meta-learner and FAM mechanism. }} \label{table:param}
    \centering
    \begin{tabular}{|@{}c@{}|c||c|c||c|}  
    \hline
        \multicolumn{5}{|c|}{Parameter Count Reduced} \\ \hline
        ~ & ~ & \multicolumn{2}{|c||}{\# Non-Zero Parameters } & ~ \\ \hline
        \textbf{S. No.\hspace{0.1em}}  & \textbf{Dataset} & \textbf{Meta-learner} & \textbf{FAM} & \textbf{Sparsity(\%)}\\ \hline
        
        1 & $CIFAR-100$ & 114373 & 23493 & 79.46 \\ \hline
        2 & $MiniImagenet$\hspace{0.15em} & 121093 & 24837 & 79.49 \\ \hline
        3 & $MRI's$ & 313986 & 94780 & 69.81 \\ \hline
        
    \end{tabular}
\end{table}

\textbf{Table \ref{table:param}} shows the count of parameters for adaptation in meta-learned and the FAM mechanism. Sparsity in percentage indicates the reduction in parameters which drastically reduces communication overhead during training in a federated setup.

We also evaluated the Vanilla Federated model performance to compare with FAM mechanism. CNN is trained in the federated setup same section of classes for ten clients. The FedAvg method \cite{pmlr-v54-mcmahan17a} is used for aggregation, and two random clients are selected for aggregation at the server.

\begin{table}[!h]
\scriptsize
    \caption{\small{Federated model $CIFAR-100$ performance. }} 
    \label{table:fed-cifar}
    \centering
    \begin{tabular}{|@{}c@{}|c||c|c|c|c|}
    \hline
        \multicolumn{6}{|c|}{Vanilla Federated - $CIFAR-100$} \\ \hline
        \textbf{S. No.\hspace{0.1em}}  & \textbf{Data per class} & \textbf{Accuracy} & \textbf{Precision} & \textbf{Recall} & \textbf{F1-score}\\ \hline
        
        1 & 5 & 53.00 & 54.53 & 53.00 & 52.81 \\ \hline
        2 & 10 & 63.40 & 64.39 & 63.40 & 63.34 \\ \hline
        3 & 20 & 65.20 & 65.59 & 65.20 & 65.28 \\ \hline
        
    \end{tabular}
\end{table}

\textbf{Table \ref{table:fed-cifar}} shows results for CNN trained in the federated setup on $CIFAR-100$ dataset sections with five distinct classes. We achieved accuracy, precision, recall and f1-score for 5 data per class is 53.00\%, 54.53\%, 53.00\% and 52.81\%, for 10 data per class is 63.40\%, 64.39\%, 63.40\% and 63.34\%, and for 20 data per class is 65.20\%, 65.59\%, 65.20\% and 65.28\%, respectively.

\begin{table}[!h]
\scriptsize
    \caption{ \small{Federated model $MiniImagenet$ performance. }} \label{table:fed-mini}
    \centering
    \begin{tabular}{|@{}c@{}|c||c|c|c|c|}
    \hline
        \multicolumn{6}{|c|}{Vanilla Federated - $MiniImagenet$} \\ \hline
        \textbf{S. No.\hspace{0.1em}}  & \textbf{Data per class} & \textbf{Accuracy} & \textbf{Precision} & \textbf{Recall} & \textbf{F1-score}\\ \hline
        
        1 & 5 & 28.40 & 37.28 & 28.40 & 25.95 \\ \hline
        2 & 10 & 38.20 & 40.89 & 38.20 & 37.36 \\ \hline
        3 & 20 & 40.20 & 37.45 & 40.20 & 37.27 \\ \hline
        
    \end{tabular}
\end{table}

\textbf{Table \ref{table:fed-mini}} shows results for CNN trained in the federated setup on $MiniImagenet$ dataset sections with five distinct classes. We achieved accuracy, precision, recall and f1-score for 5 data per class is 28.40\%, 37.28\%, 28.40\% and 25.95\%, for 10 data per class is 38.20\%, 40.89\%, 38.20\% and 37.36\%, and for 20 data per class is 40.20\%, 37.45\%, 40.20\% and 37.27\%, respectively.

\begin{table}[!h]
\scriptsize
    \caption{ \small{Federated model $Brain$-$MRI$ performance.}} \label{table:fed-brain}
    \centering
    \begin{tabular}{|@{}c@{}|c||c|c|c|c|}
    \hline
        \multicolumn{6}{|c|}{Vanilla Federated - $Brain$-$MRI$} \\ \hline
        \textbf{S. No.\hspace{0.1em}}  & \textbf{Data per class} & \textbf{Accuracy} & \textbf{Precision} & \textbf{Recall} & \textbf{F1-score}\\ \hline
        
        1 & 5 & 62.00 & 63.47 & 62.00 & 60.94 \\ \hline
        2 & 10 & 72.00 & 74.69 & 72.00 & 71.22 \\ \hline
        3 & 20 & 78.50 & 83.02 & 78.50 & 77.74 \\ \hline
        
    \end{tabular}
\end{table}

\textbf{Table \ref{table:fed-brain}} shows results for CNN trained in the federated setup on $Brain$-$MRI$ dataset sections. We achieved accuracy, precision, recall and f1-score for 5 data per class is 62.00\%, 63.47\%, 62.00\% and 60.94\%, for 10 data per class is 72.00\%, 74.69\%, 72.00\% and 71.22\%, and for 20 data per class is 78.50\%, 83.02\%, 78.50\% and 77.74\%, respectively.

\begin{table}[!h]
\scriptsize
    \caption{ \small{Federated model $Cardiac$-$MRI$ performance. }} \label{table:fed-cardiac}
    \centering
    \begin{tabular}{|@{}c@{}|c||c|c|c|c|}
    \hline
        \multicolumn{6}{|c|}{Vanilla Federated - $Cardiac$-$MRI$} \\ \hline
        \textbf{S. No.\hspace{0.1em}}  & \textbf{Data per class} & \textbf{Accuracy} & \textbf{Precision} & \textbf{Recall} & \textbf{F1-score}\\ \hline
        
        1 & 5 & 58.25 & 58.31 & 58.25 & 58.17 \\ \hline
        2 & 10 & 63.75 & 63.77 & 63.75 & 63.74 \\ \hline
        3 & 20 & 65.50 & 65.66 & 65.50 & 65.41 \\ \hline
        
    \end{tabular}
\end{table}

\textbf{Table \ref{table:fed-cardiac}} shows results for CNN trained in the federated setup on $Cardiac$-$MRI$ dataset sections. We achieved accuracy, precision, recall and f1-score for 5 data per class is 58.25\%, 58.31\%, 58.25\% and 58.17\%, for 10 data per class is 63.75\%, 63.77\%, 63.75\% and 63.74\%, and for 20 data per class is 65.50\%, 65.66\%, 65.50\% and 65.41\%, respectively.

\begin{table}[!h]
\scriptsize
    \caption{ \small{Federated model $Alzheimer$-$MRI$ performance. }} \label{table:fed-alz}
    \centering
    \begin{tabular}{|@{}c@{}|c||c|c|c|c|}
    \hline
        \multicolumn{6}{|c|}{Vanilla Federated - $Alzheimer$-$MRI$} \\ \hline
        \textbf{S. No.\hspace{0.1em}}  & \textbf{Data per class} & \textbf{Accuracy} & \textbf{Precision} & \textbf{Recall} & \textbf{F1-score}\\ \hline
        
        1 & 5 & 56.75 & 59.32 & 56.75 & 53.55 \\ \hline
        2 & 10 & 60.50 & 60.63 & 60.50 & 60.38 \\ \hline
        3 & 20 & 69.75 & 70.90 & 69.75 & 69.33 \\ \hline
        
    \end{tabular}
\end{table}

\textbf{Table \ref{table:fed-alz}} shows results for CNN trained in the federated setup on $Alzheimer$-$MRI$ dataset sections. We achieved accuracy, precision, recall and f1-score for 5 data per class is 56.75\%, 59.32\%, 56.75\% and 53.55\%, for 10 data per class is 60.50\%, 60.63\%, 60.50\% and 60.38\%, and for 20 data per class is 69.75\%, 70.90\%, 69.75\% and 69.33\%, respectively.

\begin{table}[!h]
\scriptsize
    \caption{ \small{Federated model $BreastCancer$-$MRI$ performance. }} \label{table:fed-breast}
    \centering
    \begin{tabular}{|@{}c@{}|c||c|c|c|c|}
    \hline
        \multicolumn{6}{|c|}{Vanilla Federated - $BreastCancer$-$MRI$} \\ \hline
        \textbf{S. No.\hspace{0.1em}}  & \textbf{Data per class} & \textbf{Accuracy} & \textbf{Precision} & \textbf{Recall} & \textbf{F1-score}\\ \hline
        
        1 & 5 & 76.75 & 76.78 & 76.75 & 76.74 \\ \hline
        2 & 10 & 78.75 & 79.77 & 78.75  & 78.57\\ \hline
        3 & 20 & 88.75 & 88.80 & 88.75 & 88.75 \\ \hline
        
    \end{tabular}
\end{table}

\textbf{Table \ref{table:fed-breast}} shows results for CNN trained in the federated setup on $BreastCancer$-$MRI$ dataset sections. We achieved accuracy, precision, recall and f1-score for 5 data per class is 76.75\%, 76.78\%, 76.75\% and 76.74\%, for 10 data per class is 78.75\%, 79.77\%, 78.75\% and 78.57\%, and for 20 data per class is 88.75\%, 88.80\%, 88.75\% and 88.75\%, respectively.

\begin{table}[!h]
\scriptsize
    \caption{ \small{Federated model $Prostate$-$MRI$ performance. }} \label{table:fed-prost}
    \centering
    \begin{tabular}{|@{}c@{}|c||c|c|c|c|}
    \hline
        \multicolumn{6}{|c|}{Vanilla Federated - $Prostate$-$MRI$} \\ \hline 
        \textbf{S. No.\hspace{0.1em}}  & \textbf{Data per class} & \textbf{Accuracy} & \textbf{Precision} & \textbf{Recall} & \textbf{F1-score}\\ \hline
    
        1 & 5 & 72.50 & 72.68 & 72.50 & 72.44 \\ \hline
        2 & 10 & 76.75 & 76.76 & 76.75 & 76.75 \\ \hline
        3 & 20 & 84.50 & 85.09 & 84.50 & 84.43 \\ \hline
    \end{tabular}
\end{table}

\textbf{Table \ref{table:fed-prost}} shows results for CNN trained in the federated setup on $Prostate$-$MRI$ dataset sections. We achieved accuracy, precision, recall and f1-score for 5 data per class is 72.50\%, 72.68\%, 72.50\% and 72.44\%, for 10 data per class is 76.75\%, 76.76\%, 76.75\% and 76.75\%, and for 20 data per class is 84.50\%, 85.09\%, 84.50\% and 84.43\%, respectively.

\subsection{Discussion}
The findings support the objective of the FAM mechanism, which is to learn a sparse global model and personalize it on clients. The FAM mechanism could produce a common model with learned common features that was able to adapt using a small amount of data present locally on clients. MRI images obtained at different institutions with different MRI scanners could perform well after personalization to classify various healthcare problems. 

Two benchmark datasets, $CIFAR-100$ and $MiniImagenet$, as well as healthcare-specific datasets, $Brain-MRI$, $Cardiac-MRI$, $Alzheimer-MRI$, $Breast Cancer-MRI$, and $Prostate-MRI$, were used to validate the FAM mechanism. Compared to Vanilla FL, our experiments advocated that the meta-learned without pruning and the FAM mechanism would perform better after adaptation on fewer samples. Meanwhile, the FAM mechanism demonstrated superior adaptability concerning meta-learned adaptation under identical configurations. Additionally, the FAM mechanism effectively minimized communication overhead through substantial parameter pruning with an appreciable increase in accuracy.

\section{Conclusion}
In this work, we focused on the challenge of data insufficiency on clients with data heterogeneity-induced domain shift, which necessitates the personalization of the learnt global model. A Fast Adaptive Federated Meta-learning mechanism was proposed to meet the challenge of efficient training a global model in the federated learning setting with adaptation using client-specific data. We exploited the LTH approach in the federated meta-learning setting to train a sparse global that learns common features across clients. Personalization was achieved using MAML-based meta-learning on clients. The results for healthcare-specific datasets along with two benchmark datasets, showed that the globally shared meta-learner could learn the common features among all datasets, which was able to adapt quickly with very few data even under the presence of domain shift in the data. The accuracy of the sparse model was slightly better, with a large percentage of parameter pruning, so the communication overhead was reduced significantly. The adaptation through model growth led to fast adaptation and increased accuracy. This confirmed the effectiveness and efficiency of the FAM mechanism.

\bibliographystyle{unsrt}  
\bibliography{references}

\end{document}